\documentclass[submission,copyright,creativecommons]{eptcs}
\providecommand{\event}{WIVACE 2013} 
\usepackage{breakurl}             
\usepackage{algpseudocode}
\usepackage{algorithm}
\usepackage{amssymb}
\usepackage{amsmath}
\usepackage{graphicx}
\title{A Hybrid Monte Carlo Ant Colony Optimization Approach for Protein Structure Prediction in the HP Model}
\author{Andrea G. Citrolo $\quad$ Giancarlo Mauri
\institute{Dipartimento di Informatica Sistemistica e Comunicazione\\
Universit\`a degli Studi di Milano-Bicocca.}\\
\email{\{citrolo,mauri\}@disco.unimib.it}
}
\def\titlerunning{ HMCACO for Protein Structure Prediction in the HP Model}
\def\authorrunning{A. G. Citrolo,  G. Mauri}
\begin{document}
\maketitle

\begin{abstract}
The hydrophobic-polar model has been widely studied in the field of protein structure prediction 
both for theoretical purposes and as a benchmark for new optimization strategies.  In this work we introduce a new 
heuristics based on Ant Colony Optimization and Markov Chain Monte Carlo that we called Hybrid Monte Carlo Ant Colony 
Optimization. We describe this method and compare results obtained on well known HP instances in the 3-dimensional
cubic lattice to those obtained with standard Ant Colony optimization and Simulated Annealing. All methods were implemented using an unconstrained neighborhood 
and a modified objective function to prevent the creation of overlapping walks. Results show that our methods perform 
better than the other heuristics in all benchmark instances.

\end{abstract}

\section{Introduction}

Proteins are polymers composed of linear chains of amino-acids. These molecules adopt a complex three-dimensional structure that allows 
them to take part in almost every biological process. Since proteins function strongly depends on their three-dimensional 
structure the availability of fast and reliable techniques for protein structure determination is of great interest in 
fields such as drug design, proteomics and synthetic biology. Nevertheless, experimental techniques for protein structure 
determination, like X-ray crystallography and nuclear magnetic resonance are still very expensive and time consuming.
With the increasing number of sequence data available, the lack of a reliable method for protein structure prediction (PSP)
risks to become the true 
bottleneck in the post-genomic era, making PSP one of the most challenging problems in computational biology. 
According to the widely accepted thermodynamic hypothesis \cite{Anfinsen1973}, at least for small globular proteins, the native structure
is the structure with the lowest potential energy and so the PSP can be treated as a global minimization problem. 
Real instances of PSP contain thousand of atoms independently moving in the three-dimensional space, so 
they are very hard to treat both in terms of computational resources and in terms of theoretical formalization. For these
reasons some simplified models have been introduced in order to investigate fundamental aspects of the folding process.
Among them the most studied is the hydrophobic-polar (HP) model introduced by Dill et al. \cite{Dill1985} in which proteins are represented as strings
in a binary alphabet; the goal is to find the optimal on-lattice 
self-avoiding walk for the given binary sequence. Each protein residue in the HP model is  represented by a single position
in the input string. 
The objective function counts the contacts between non-adjacent hydrophobic positions in the sequence. 
This model has been studied in a variety of lattices, and it has been proved that the optimization problem of maximizing
the number of H-H contacts is NP-hard for a broad class of objective functions independently of the lattice chosen to define
the discrete protein-folding model \cite{Hart1997}. In recent years a great number of methods have been applied to the PSP in the HP model,
ranging from constraint programming to heuristic approaches \cite{Albrecht2008,Hsu2003,Hsu2011,Liang2001,Mann2008,Piccolboni1998,Shatabda2013,Shmygelska2005,Thachuk2007,Ullah2010}.
Analyzing these studies two considerations can be made: primary
methods based on perturbation of complete solutions (like Replica Exchange Monte Carlo) seem to perform better than 
methods based on constructive approaches (with the notable exception of the pruned-enriched Rosenbluth method). 
In addition methods that keep memory of good solutions to prevent re-sampling performs better than methods that do not \cite{Shatabda2013}. 
In this work we introduce a new heuristics that combines Ant Colony Optimization (ACO) with the Monte Carlo method in order to 
overcome major drawbacks of these techniques.

\section{Solution Representation}
\begin{figure}
\centering
\includegraphics[width = 6 cm]{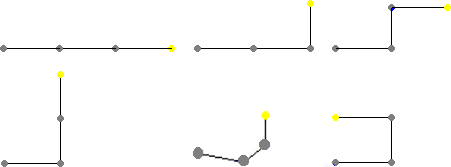}
\caption{Allowed starting blocks}
\label{fig:starting_blocks}
\end{figure}

In a three-dimensional cubic lattice residue positions are encoded using three integer numbers representing 
respectively values along x, y and z Cartesian axis. In the literature solutions coding  for a generic HP string $s$  
of length $n$ are generally represented as strings of length $n-1$, in which each position can assume one of six values
indicating an unitary movement along one of the six directions. The first position is arbitrarily placed at the origin
of the axes. In this work we use a slightly different solution representation that is similar to the one adopted in off-lattice
models based on the fragment assembly strategy \cite{Xu2012}. We build 150 fragments that cover all the valid conformations of an HP string
of length four. Each fragment determines the local structure of a triplet of residues 
starting at the origin and also the position of the first residue in the following triplet. 
To prevent the creation of rotational duplicates only the six blocks in fig~.\ref{fig:starting_blocks} 
are allowed at the first position.

\section{Objective Function}
The canonical objective function of the HP problem considers only the number of contacts between 
non consecutive hydrophobic positions (H-H contacts) and the problem is set as a global minimization. 
As we will discuss later, in this work we use a simple neighborhood structure based on blocks exchange.
This Neighborhood does not guarantee that generated solutions are self-avoiding walks. In order to
discourage the generation of overlapping walks and to improve the performance of the constructive step we 
introduced the modified scoring function shown in eq.~\ref{eq:Cost}. 

\begin{equation}
 \label{eq:Cost}
 Energy = -[({HHscore \over 1 + OVscore})^2 + {1 \over 1+ HPscore} + {2 tHH\over 1 + HDscore}]
\end{equation}

where $tHH$ is the upper bound of number of contacts defined in \cite{Istrail2009}. The general form of a $score$ is defined above:

\begin{equation}
 \label{eq:Score}
score = \sum_{i = 1}^{n-3} \sum_{j = i + 3}^{n} f(\chi_{i}, \chi_{j}, s_{i}, s_{j}) 
\\
   \\
    \quad i,j \in \mathbb{N} 
\end{equation}

where  $\chi_i, \chi_j \in \mathbb{Z}^3$ are the vectors of coordinates of residue i,j respectively 
while $s_i, s_j \in [0,1]$ are values of sequence at position i,j 
(we arbitrarily choose 1 for hydrophobic positions and 0 for polar positions), $f$ is a function that defines 
the specific type of score.
The $HHscore$ represents the canonical objective function of the HP problem and is obtained setting $f=\kappa$ 
as defined in in eq.~\ref{eq:HH} :

\begin{equation}
 \label{eq:HH}
 \kappa(\chi_i, \chi_j, s_i, s_j) =
  \begin{cases}
   -1 & \text{if } \sum_{k = 1}^{3} |\chi_{i,k}-\chi_{j,k}| = 1 \land s_i = s_j = 1 \\
   0       & \text{otherwise}
   \end{cases}    
\end{equation}

$OVscore$ uses $f=\omega$ shown in eq.~\ref{eq:OV}; it counts the number of overlapping positions. To also consider the
overlaps at the boundary between blocks the $\omega$ function is also computed between residues i and i + 1 at 
the boundary positions.

\begin{equation}
 \label{eq:OV}
 \omega(\chi_i, \chi_j, s_i, s_j) =
  \begin{cases}
   1 & \text{if } \sum_{k = 1}^{3} |\chi_{i,k}-\chi_{j,k}| = 0 \\
   0       & \text{otherwise}
   \end{cases} 
\end{equation}

$HPscore$ uses $f=\phi$ shown in eq.~\ref{eq:HP} ; it counts the contacts between hydrophobic and polar positions (H-P contacts),
it was introduced to prevent the formation of undesired H-P contacts; a similar term was used in \cite{Hsu2003}.  
 
 \begin{equation}
 \label{eq:HP}
 \phi(\chi_i, \chi_j, s_i, s_j) =
  \begin{cases}
   1 & \text{if } \sum_{k = 1}^{3} |\chi_{i,k}-\chi_{j,k}| =1 \land s_i \neq s_j \\
   0       & \text{otherwise}
   \end{cases} 
\end{equation}

finally the $HDscore$ uses $f=\delta$ in eq.~\ref{eq:HD} and it was introduced to bias the early steps of 
constructive methods toward compact solutions.

\begin{equation}
\label{eq:HD}
  \delta(\chi_i, \chi_j, s_i, s_j) =
  \begin{cases}
   \sum_{k = 1}^{3} |\chi_{i,k}-\chi_{j,k}| & \text{if $i$, $j$ have different parity } \land s_i = s_j = 1 \\
   0       & \text{otherwise}
   \end{cases} 
\end{equation}

\section{Graph mapping}
Since both ACO heuristics and Hybrid Monte Carlo Ant Colony Optimization (HMCACO) require an explicit probabilistic model 
that influences transitions between states of the modeled system a graph mapping of PSP problem is needed.
Given an HP string $s$ of length $n$, we define
$b_s \in \mathbb{N}$ to be the size of a structural block in terms of sequence coverage ($b_s = 3$ in our representation),
$n_b \in \mathbb{N}$ to be the number of structural blocks used to represent each sub-sequence of size $b_s$\; ($n_b = 150$).
Furthermore we define:

\begin{equation}
\label{eq:GR}
  ID = \{i \in \mathbb{N} | i \leq n/b_s\}, \quad
  LS = \{m \text{ is a valid local structure of size} b_s\}, \quad
  A = ID \times LS
\end{equation}

an undirected graph $G$ of size $(n_b{n\over b_s})$ such that there is a one to one correspondence between nodes of
$G$ and elements of $A$.
In the general case the set of edges of $G$ contains one edge between any pair of node in $i$,$j$ 
for which  holds $id_i \neq id_j$ in the complementary representation. 
Finally, we define $G'$ that contains the same node set of $G$
but with edges connecting only pairs of nodes for which it holds  $|id_i - id_j| = 1$ and oriented in the direction 
of increasing values of $id$. 
The HP problem can be formalized as a search for paths of size ${n\over b_s}$ in a dynamical 
graph dynamical system (GDS) defined on $G'$ in which the state of the system  is 
obtained through a backtracking function that builds the coordinates matrix corresponding to the 
structure encoded by the nodes of the path.
Since each path on $G'$ generates a state that is a valid argument for
the evaluation function in eq~.\ref{eq:Cost}, the PSP problem is that of finding the lowest energy path. 
As we will see, both ACO and HMCACO use the $G$ graph to keep track of good pairwise relations between structural
blocks obtained by lower energy paths sampled by ants in the DGS defined on $G'$.  This information is then
used to bias the following samplings of the GDS.

\section{Methods}
\subsection{Hill Climbing}
This simple iterative heuristics is based on the concept of neighborhood. The neighborhood is built through the definition
of a neighborhood function of the following type:
\begin{center}
 Let  I  be an instance of an optimization problem \\
 Let  X  be the set of all feasible solutions for  I \\
 Let  $H_{s}$  be the set of neighbor solutions of   $s$, \\ $s,\in X$, $H_{s}\subseteq X$ \\ 
 we define: \\
 $N: X\rightarrow 2^{X}: \quad \forall s \in X \rightarrow N(s) \in H_{s}$ \\
\end{center}
In this work we state that a solution $a$ is k-neighbor of a solution $b$  if and only if they differs
for less than k structural block. 
The hill climbing procedure used in this work performs at each iteration
an exhaustive search over the  1-neighborhood of the input solution.
The best neighbor found according to eq~.\ref{eq:Cost}
is thus selected as the new current solution only if its cost is lower than the cost of the input solution.
This heuristics was used as local optimizer in both ACO and HMCACO.

\subsection{Simulated Annealing}
Here we give just an idea of the basic concept underlying the Simulated Annealing method, for an extensive treatment
of this topic see \cite{Url2007,Cerny1985}.
Simulated Annealing (SA) is one of the most studied neighborhood based heuristics
of the broad Markov Chain Monte Carlo (MCMC) family of methods.  
Many heuristics in this family have proved to perform very well on HP instances when the neighborhood is obtained 
through pool moves \cite{Lesh2003} or with the move set described in \cite{Gurler1983}.
A general Markov Chain Monte Carlo method works in an iterative fashion 
sampling the search space according to a given distribution. 
The Boltzmann distribution is the most used one for polymer simulation.
This means that most of the sampling time is spent on solutions with low energy values.
Each iteration of a MCMC method can be divided into two phases: sampling, and transition.  
In the sampling phase a random neighbor of the current solution is generated,
then in the acceptance phase the Metropolis Hastings acceptance criterion in eq~.\ref{eq:SA1} is
applied, and the result is compared to a random number between [0,1). 
If p is greater than the random number, the
neighbor solution is accepted as starting solution of the next iteration.

\begin{equation}
 \label{eq:SA1}
 p(s, s') =
  \begin{cases}
   1 & \text{if } c_s' \leq c_s  \\
   exp^{{c_s - c_s'} \over T} & \text{otherwise}
   \end{cases}    
\end{equation}

Here $p(s, s')$ is the probability to make the transition from solution $s$ to neighbor solution $s'$, while
$c_i$ is the value of the objective function of solution $i$.
The peculiarity of SA
is that the temperature varies during the search process. At the beginning of the search high values of temperature are
used to facilitate a broad exploration of the search space, after each iteration the temperature is decreased using a 
problem-dependent cooling scheme. In this work we used for SA the same neighborhood function
presented for Hill Climbing and we based the cooling on the scheme presented in \cite{Albrecht2008}. 

\subsection{Ant Colony Optimization}
Ant colony optimization is a bio-inspired meta-heuristics to approach hard combinatorial problems in which a colony of
simple agents (artificial ants) interact to efficiently explore the search space.
The only applicability 
condition to satisfy in order to use ACO is the availability of a graph mapping for the transitions in the
target problem\cite{M.2004}. As we explained before, transitions of the PSP problem can be mapped on a graph.
The general idea of ant inspired systems is that of combining the constructive strategy with a global evaluation
stochastic heuristics and to keep memory of the relations between solution components in high quality solutions.
This relation is stored in a matrix called pheromone matrix that is a real number representation of the edges 
of the previously described transition graph.
In this work we adopted the  $Max$-$Min$ Ant System meta-heuristics \cite{Stutzle2000}; 
this implies that the quantity of pheromone in each position
of the pheromone matrix is bounded by a minimal and a maximal value.
Each iteration of the meta-heuristics is composed of three main steps: construction, evaluation and daemon actions. 
The construction step works like a probabilistic greedy algorithm. A starting node 
with $id = 1$ is chosen probabilistically according to the mean of pheromone values associated to its edges
(in case of PSP only nodes with $id=1$ are considered as candidate starting nodes).
All the subsequent nodes that will enter the solution are chosen in probability as follows:
\begin{equation}
 \label{eq:ACO1}
 p(i,j) = {t_{ij}^{\alpha}\; \eta^{\beta} \over \sum_{b = 1}^{n_b}t_{ib}^{\alpha} \; \eta_{c}^{\beta}}
\end{equation}
Here $p(i,j)$ is the probability of including node $j$ in the growing solution, $t_{ij}$ is the pheromone value on
the $ij$-edge $\eta$ is the cost of extending the current partial solution with node $j$, computed 
using a heuristic function, $\alpha$ and $\beta$ are algorithm parameters.
Since ants build solutions in a constructive fashion, the pheromone matrix for ACO uses only edge between
consecutive block in the sequence so in this case the edge set of $G$ coincides with that of  $G'$.
In the daemon-step, solutions that have been built from the ants undergo local optimization. Then the pheromone level 
is decreased for all the edges according to eq~.\ref{eq:ACO2}.
\begin{equation}
\label{eq:ACO2}
 \tilde{t}_{ij} = (1-\rho) t_{ij} + \rho t_{min} 
\end{equation}
Here $\bar{t}_{ij}$ is the pheromone value on $ij$-edge after evaporation, while $\rho$ is a parameter.
In the evaluation step the best solution of the current iteration is
compared to the best solution obtained so far. If current solution is an improvement over the
best solution, the latter is updated and then used to increase the pheromone level on the contained edges, 
as shown in eq.~\ref{eq:ACO3}. Otherwise, the releaser is chosen using some heuristic criterion: 

\begin{equation}
\label{eq:ACO3}
\hat{t_{ij}} = 
\begin{cases}
\tilde{t}_{ij}+{c_{s}\over c_{opt}^*} & \text{if } \tilde{t}_{ij} < t_{max} \\
   t_{max}       & \text{otherwise}
\end{cases}
\end{equation}
Here $\hat{t_{ij}}$ is the value of pheromone matrix at positions $i$, $j$ after the release, while
$c_s$ is the cost of the releaser solution and $c_{opt}$ is an estimate of the optimal value for the given instance. 
At the beginning of the algorithm the whole pheromone matrix is set to the max pheromone value. 
In this work we used five ants for each iteration; we set $\alpha = \beta = 1$, $\rho = 0.1 $.  
The policy to set and to update values of  $t_M$ and $t_m$ has been defined following suggestions from \cite{M.2004}. 

\subsection{Hybrid Monte Carlo Ant Colony Optimization}
As anticipated in the introduction, the technique we present here is aimed to combine
the pheromone biased search typical of ACO with the perturbation
approach of a Markov Chain Monte Carlo method. The general structure of the algorithm is that
of an Ant based heuristics already described. The main difference
with ACO is that in the construction step each ant is initialized to a model solution, and a set of 
pheromone based perturbations is applied. This introduces the concept of neighborhood typical of a
perturbation based approach. During perturbation 
each candidate node is evaluated using the heuristic information and the mean of the pheromone values computed 
over all edges connecting the incoming node to nodes in the solution that are not going to be replaced.
The probability to select a generic incoming node thus becomes:
\begin{equation}
 \label{eq:HMCACO1}
 p(i,j) = {
 ({1\over |S|}
 \sum_{\sigma = 1}^{|S|}
 t_{ij}^{\alpha})\; \eta^{\beta} 
 \over
 \sum_{b = 1}^{n_b} ({1\over |S|}\sum_{\sigma = 1}^{|S|}t_{\sigma b}^{\alpha}) \; \eta_{c}^{\beta}}
\end{equation}
Here p(i,j) is the probability to accept node j as perturbing node at position $i$; $S$ is the set containing  nodes in the ant 
that are not going to be replaced by node $j$, so $|S| = {n \over b_s}-1$, and $\sigma$ is an index over $S$.
It is clear from eq~.\ref{eq:HMCACO1} that all the edges of $G$ are considered in the pheromone matrix of HMCACO.
A second modification introduced in HMCACO with respect to the standard ACO is that comparison between 
best so far solution and iteration best solution
is made using the Metropolis Hastings criterion eq~.\ref{eq:SA1}. The accepted solution is allowed both to release the 
pheromone on contained edges and to become the model for the next iteration. The values of the parameters for HMCACO are the 
same described for ACO; the optimal value for neighborhood size was found to be four.

\section{Results and Discussion}
In this work we chose the standard benchmark set for HP model in the three-dimensional cubic lattice taken from \cite{Xz} and shown
in tab.1 .

\begin{table}
\begin{center}
\label{tab:1}
\begin{tabular}{|l|c|}
\hline
 ID & sequence\\
 \hline
 S1 & hphhpphhhhphhhpphhpphphhhphphhpphhppphpppppppphh\\
 S2 & hhhhphhphhhhhpphpphhpphpppppphpphppphpphhpphhhph\\
 S3 & phphhphhhhhhpphphpphphhphphppphpphhpphhpphphpphp\\
 S4 & phphhpphphhhpphhphhppphhhhhpphphhphphpppphpphphp\\
 S5 & pphppphphhhhpphhhhphhphhhpphphphpphpppppphhphhph\\
 S6 & hhhppphhphphhphhphhphppppppphphpphppphpphhhhhhph\\
 S7 & phpppphphhhphphhhhphhphhppphphppphhhpphhpphhppph\\
 S8 & phhphhhphhhhpphhhpppppphphhpphhphppphhphphphhppp\\
 S9 & phphpppphphphpphphhhhhhpphhhphpphphhpphphhhpppph\\
 S10& phhpppppphhppphhhphpphphhpphpphpphhpphhhhhhhpphh\\
 \hline
\end{tabular}
\caption{Standard benchmark sequences for 3D HP problem in cubic lattice}
\end{center} 
\end{table}

This set has been extensively used in other works \cite{Albrecht2008,Liang2001,Shmygelska2005,Thachuk2007}. All the sequences in the benchmark set have length 48, and the global 
optimum for each of them has been obtained using the CPSP tool \cite{Mann2008}. This program uses a constraint based approach to perform an 
exhaustive search for sequences of moderate size. All the methods have been implemented using the same library and tested 
on the same hardware, so we based the comparison on results obtained from short run of comparable length (CPU time $< 2'$). 
In table tab.2 we show the best and average results for each instance computed over 50 runs
for the different heuristics. None of the methods was able to reach the global optimum in the considered execution time.
This is probably due to the availability of overlapping solutions introduced from the naive neighborhood structure we used.
It is very likely that the presence of these solution alters the fitness landscape of the problem making it rougher. 
An argumentation in support of this supposition is that in the work of Albrecht et al. \cite{Albrecht2008} SA with a neighborhood based on
pull moves is able to reach the global optimum for several instances in a number of iterations lower than what we used 
in this work.
In the case of ACO, previously reported results \cite{Shmygelska2005} indicate that also with other neighborhood structures the time
required to reach the global optimum is longer than the time used in this work. The poor performances obtained from SA
in this context however are not completely unexpected, if we consider the difference in size
between a pull move based neighborhood and the one adopted in this work.
Albrecht et al. in their study of time complexity of HP model for SA had shown that the number of iterations required 
to reach the global minimum in canonical HP structure prediction is bounded to:

\begin{equation}
 t_{opt} = ({m \over \delta})^{D/\gamma}
\end{equation}

Here $t_{opt}$ is the number of iterations required to reach the global optimum of the problem with a confidence of $\delta$, 
$m$ is the average size of the neighborhood of a solution,
$D$ is the depth in terms of energy of the deepest local minimum for the given instance and $\gamma$ is the 
average energy variation in an improving iteration. 
Even if it is hard to compare convergence properties of different models, we can assume that the size of the 
neighborhood has similar effects on the general behavior of SA, and this could explain the bad performance we observed. 
In the work of Albrecht, in fact, the value of $m$ was estimated to be
$m\approx n/2$, while in this work it is equal to the number of blocks allowed in each position $m = 150$.
Another interesting observation is that ACO in this context performs better than SA. This is probably due to 
the pheromone bias that helps the method to avoid overlapping solutions and to
spend more time on feasible solutions. This is true also for HMCACO that is able to sample low energy
regions in the search space with increased efficacy. In our opinion this is an effect of the model-based
perturbation strategy, since  
in standard ACO the constructive approach, even in the case of a single unlucky insertion, can push the search
to regions of space with low pheromone content, impairing also the following insertion steps.

\begin{table}
\label{tab:2}
 \begin{center}
\begin{tabular}{|l|*{4}{c|}}
\hline
 ID & $E_{min}$ & SA & ACO & HMCACO\\
 \hline
 S1 & -32 & \textbf{-26}(-23.7) & \textbf{-29}(-27.0) & \textbf{-31} (-28.6)\\
 S2 & -34 & \textbf{-27}(-23.9) & \textbf{-29}(-26.7) & \textbf{-32} (-29.2)\\
 S3 & -34 & \textbf{-28}(-25.1) & \textbf{-28}(-26.7) & \textbf{-32} (-29.5)\\
 S4 & -33 & \textbf{-28}(-24.1) & \textbf{-29}(-26.6) & \textbf{-31} (-29.1)\\
 S5 & -32 & \textbf{-28}(-25.1) & \textbf{-29}(-26.5) & \textbf{-31} (-28.5)\\
 S6 & -32 & \textbf{-24}(-23.0) & \textbf{-28}(-25.7) & \textbf{-30} (-27.7)\\
 S7 & -32 & \textbf{-26}(-23.6) & \textbf{-28}(-26.5) & \textbf{-31} (-28.4)\\
 S8 & -31 & \textbf{-27}(-24.1) & \textbf{-29}(-25.9) & \textbf{-29} (-27.8)\\
 S9 & -34 & \textbf{-28}(-25.2) & \textbf{-30}(-27.8) & \textbf{-32} (-29.7)\\
 S10& -33 & \textbf{-27}(-24.3) & \textbf{-29}(-27.0) & \textbf{-31} (-29.6)\\
 \hline
\end{tabular}
\caption{Results of different heuristics compared to optimal values of energy for each benchmark sequence}
\end{center}
\end{table}

\
\section{Conclusions and Future Works}
In this work we presented a new heuristics based on ACO and Markov Chain Monte Carlo that we called HMCACO; we tested it 
on standard benchmarks of PSP in 3D-HP model with a naive neighborhood and a modified objective function. 
Results showed that in this context HMCACO outperforms both ACO and SA.
Preliminary analysis of SA simulations indicate that neighborhood introduced in this work 
might not be well suited for the SA heuristics. This is interesting since for many aspects, the representation used here 
is closer to off-lattice model than the standard HP representation. Future work will be dedicated to establishing the 
efficacy of HMCACO for PSP both in HP models with the standard neighborhood and objective function and also in off-lattice models.

\nocite{*}
\bibliographystyle{eptcs}
\bibliography{HMCACO}
\end{document}